\documentclass{article} 
\usepackage{samples/iclr2020_conference}
\usepackage{times}

\iclrfinalcopy 

\usepackage[utf8]{inputenc}
\usepackage[english]{babel}
\usepackage[T1]{fontenc}
\usepackage{mathptmx}

\usepackage{microtype}      
\usepackage{url}
\usepackage[colorlinks=true,linkcolor=blue, citecolor=blue, filecolor=magenta, urlcolor= red,]{hyperref}
\usepackage{bookmark}
\usepackage{graphicx,grffile,subcaption,wrapfig}
\usepackage{multirow,booktabs}

\newcommand{\eg}{\textit{e.g.}}
\newcommand{\ie}{\textit{i.e.}}
\newcommand{\etc}{\textit{etc.}}
\newcommand{\nameofproject}{MagicMix}
\usepackage{amsmath,amsfonts, mathtools, bm, bbm, nicefrac}
\newcommand{\ptheta}{p_{\theta}}

\newcommand\blfootnote[1]{%
	\begingroup
	\renewcommand\thefootnote{}\footnote{#1}%
	\addtocounter{footnote}{-1}%
	\endgroup}

\newcommand{\mcal}[1]{\mathcal{#1}}

\usepackage[ruled,noline,linesnumbered]{algorithm2e}
\IncMargin{1em}
\SetKwInOut{Input}{Input}
\SetKwInOut{Output}{Output}

\newcommand{\KL}{D_{\mathrm{KL}}}

\def\vzero{{\bm{0}}}

\def\vmu{{\bm{\mu}}}

\def\vx{{\bm{x}}}
\def\vy{{\bm{y}}}

\def\mI{{\bm{I}}}

\def\mM{{\bm{M}}}

\DeclareMathAlphabet{\mathsfit}{\encodingdefault}{\sfdefault}{m}{sl}
\SetMathAlphabet{\mathsfit}{bold}{\encodingdefault}{\sfdefault}{bx}{n}

\def\sD{{\mathbb{D}}}

\def\sR{{\mathbb{R}}}

\title{\nameofproject: Semantic Mixing with Diffusion Models}
\author{Jun Hao Liew*, Hanshu Yan*, Daquan Zhou \& Jiashi Feng\\
ByteDance Inc.\\
\texttt{\{junhao.liew, hanshu.yan, daquanzhou, jshfeng\}@bytedance.com}\\
}

\begin{document}

\makeatletter
\g@addto@macro\@maketitle{
  \begin{figure}[h]
  \setlength{\linewidth}{\textwidth}
  \setlength{\hsize}{\textwidth}
  \centering
  \vspace{-2em}
  \includegraphics[width=\textwidth]
  {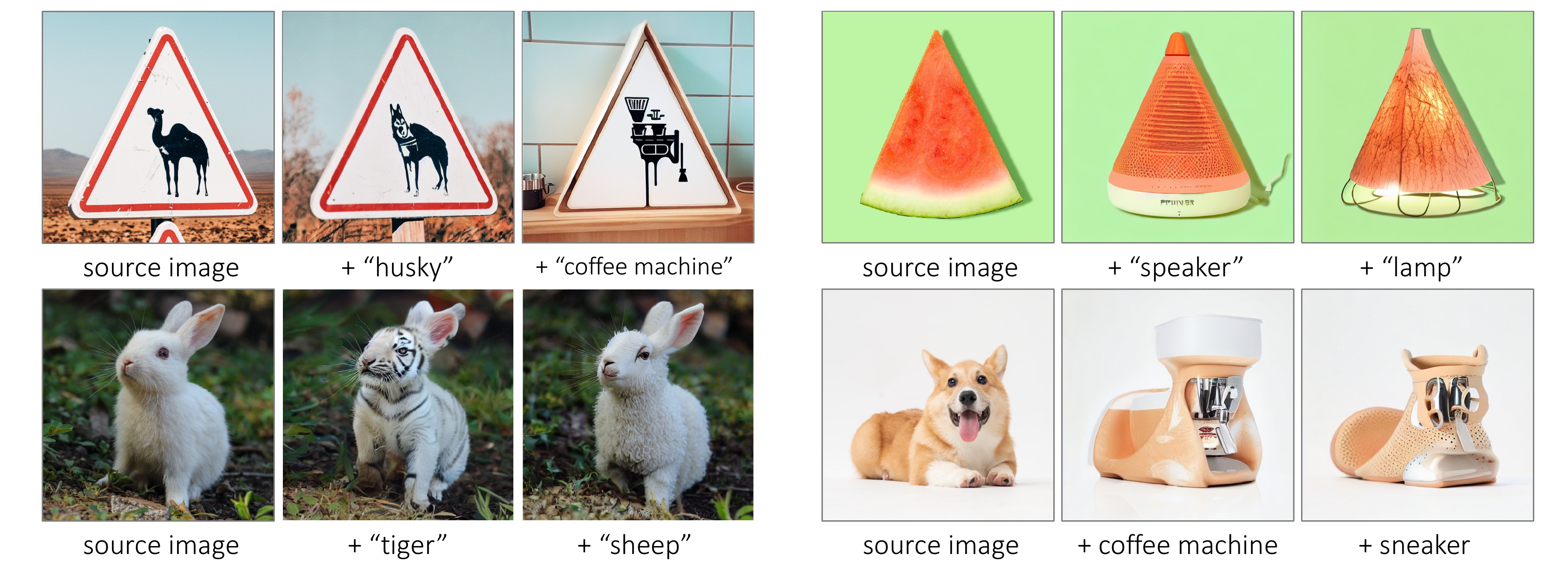}
  \caption{\textbf{\nameofproject} allows mixing of two different semantics (\eg, corgi and coffee machine) to create a novel concept (\eg, corgi-alike coffee machine). Image credit (source images): Unsplash.
  }
  \label{fig:teaser}
  \end{figure}
}
\makeatother
  
\maketitle

\begin{abstract}
Have you ever imagined what a \textit{corgi-alike coffee machine} or a \textit{tiger-alike rabbit} would look like? 
In this work, we attempt to answer these questions by exploring a new task called \textbf{semantic mixing}, aiming at blending two different semantics to create a new concept (\eg, corgi $+$ coffee machine $\rightarrow$ corgi-alike coffee machine). 
Unlike style transfer where an image is stylized according to the reference style without changing the image content, semantic blending mixes two different concepts in a semantic manner to synthesize a novel concept while preserving the spatial layout and geometry. 
To this end, we present \textbf{\nameofproject}, a simple yet effective solution based on pre-trained text-conditioned diffusion models. 
Motivated by the progressive generation property of diffusion models where layout/shape emerges at early denoising steps while semantically meaningful details appear at later steps during the denoising process, our method first obtains a coarse layout (either by corrupting an image or denoising from a pure Gaussian noise given a text prompt), 
followed by injection of conditional prompt for semantic mixing. 
Our method does not require any spatial mask or re-training, yet is able to synthesize novel objects with high fidelity. 
To improve the mixing quality, we further devise two simple strategies to provide better control and flexibility over the synthesized content.
With our method, we present our results over diverse downstream applications, including semantic style transfer, novel object synthesis, breed mixing, and concept removal, demonstrating the flexibility of our method. More results can be found on the project page \url{https://magicmix.github.io/}.
~\blfootnote{*Equal contribution}

\end{abstract}
\section{Introduction}

\begin{figure*}
    \includegraphics[width=\textwidth]{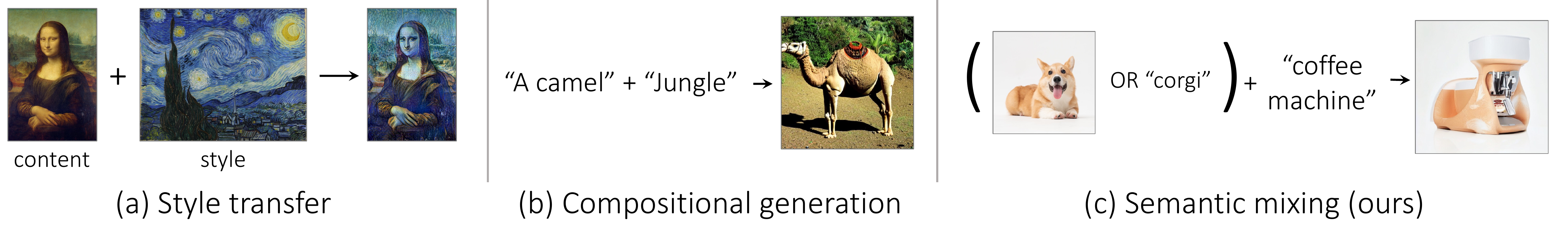}
    \caption{\textbf{Task comparison.} (a) Style transfer stylizes a content image according to the given style (\eg, The Starry Night by Vincent van Gogh, 1889) while keeping the image content (\eg, Mona Lisa by Leonardo da Vinci, 1503) unchanged. (b) Compositional generation composes multiple individual components (\eg, ``camel'', ``jungle'') to generate a complex scene. While the composition itself may be novel, each individual component is known.
    (c) Differently, semantic mixing aims to blend multiple semantics into \textit{one single} object (\eg, ``corgi'' $+$ ``coffee machine'' $\rightarrow$ ``corgi-alike coffee machine''). Image credit (input corgi image): Unsplash.}
    \label{fig:task_difference}
\end{figure*}

Have you ever imagined what a \textit{corgi-alike coffee machine} would look like? What about \textit{a rabbit that looks like a tiger}? 
Rendering such imaginary scenes is extremely challenging due to the non-existence of such objects in the real world. 
In this work, we are interested in studying a new problem termed {\textbf{semantic mixing}}, whose objective is to blend two different semantics (\eg, ``corgi'' and ``coffee machine'') in a semantic manner to create a new concept (\eg, a corgi-alike coffee machine) while being photo-realistic.

Recently developed large-scale text-conditioned image generation models, such as DALL-E 2~\citep{ramesh_hierarchical_2022}, Imagen~\citep{saharia_photorealistic_2022}, Parti \citep{yu_scaling_2022}, \etc, have demonstrated the capabilities in generating astonishing high-quality images given only text descriptions. 
Such models can even generate novel compositions (\eg, an astronaut riding a horse) due to the strong semantic prior learned from a large collection of image-caption pairs. 
Despite the novel combination, each object instance (\eg, ``astronaut'', ``horse'') is known given the learned priors. 
Besides, unlike the compositional generation (\eg, a corgi sitting beside a coffee machine), we are interested in synthesizing a novel concept (\eg, a corgi-alike coffee machine or vice-versa) by semantically mixing two different concepts. 
Nevertheless, such a problem is challenging since even a human user might not know how is it supposed to look like.

To address this, we present a new approach termed {\textbf{\nameofproject}}, which is built upon existing text-conditioned image diffusion-based generative models. Our approach is extremely simple, requiring neither re-training nor user-provided masks. 
Our method is motivated by the progressive property of diffusion-based models where layout/shape/color emerges first at early denoising steps while semantically meaningful contents appear much later during the denoising process.
Given this, we factorize the semantic mixing task into two stages: (1) layout (\eg, shape and color) semantics and (2) content semantics (\eg, the semantic category) generation. 
Specifically, consider the example of mixing ``corgi'' and ``coffee machine'', our \nameofproject\ first obtains a coarse layout semantic either by corrupting a given real photo of corgi or denoising from a pure Gaussian noise given a text prompt ``a photo of corgi''. Then, it injects a new concept (``coffee machine'' in this case) and continues the denoising process until we obtain the final synthesized results. Such a simple approach works surprisingly well in general. To improve the blending, we further devise two simple strategies to provide better control and flexibility over the generated content.

Semantic mixing is conceptually different from other image editing and generation tasks, such as the style transfer or compositional generation. Style transfer stylizes a content image according to the given style (\eg, van Gogh's The Starry Night) while preserving the image content. Compositional generation, on the other hand, composes multiple individual components to generate a complex scene (\eg, compositing ``camel'' and ``jungle'' leads to an image of a camel standing in a jungle). While the composition itself may be novel, each individual component has been already known (\ie, what does a camel look like).
Differently, semantic mixing aims to fuse multiple semantics into \textit{one single} novel object/concept (\eg, ``corgi'' $+$ ``coffee machine'' $\rightarrow$ a corgi-alike coffee machine). The differences between these tasks are illustrated in Figure~\ref{fig:task_difference}.

Thanks to the strong capability in generating novel concepts, our \nameofproject\ supports a large variety of creative applications, including semantic style transfer (\eg, generating a new sign given a reference sign layout and a certain desired content), novel object synthesis (\eg, generating a lamp that looks like a watermelon slice), breed mixing (\eg, generating a new species by mixing ``rabbit'' and ``tiger'') and concept removal (\eg, synthesizing a non-orange object that looks like an orange). Although the solution is simple, it paves a new direction in the computational graphics field and provides new possibilities for AI-aided designs for artists in a wide field, such as entertainment, cinematography, and CG effects. 

In summary, our contributions in this work are:
\begin{itemize}
    \item A new problem: semantic mixing. The goal is to synthesize a novel concept by mixing two different semantics while being photo-realistic.
    \item A new technique: MagicMix. It is built upon large-scale pre-trained text-to-image diffusion-based generative models and factorizes the semantic mixing task into the layout and content generation stage.
    \item We demonstrate several creative applications given our \nameofproject, including semantic style transfer, novel object synthesis, breed mixing, and concept removal.
\end{itemize}
\section{Related Works}
\subsection{Diffusion Probabilistic Models} The Diffusion Probabilistic Models (DPM) family has achieved great success in both unconditional and conditional generative modeling tasks \citep{ho_denoising_2020,song_denoising_2022,ho_video_2022,song_score-based_2021}, including image/video generation\citep{ho_video_2022, nichol_improved_nodate}, molecular generation\citep{xu_geodiff_2022}, and time-series modeling\citep{rasul_autoregressive_2021}. 
They are not only able to generate perceptually high-quality samples but also can yield outstanding log-likelihood scores. However, the computational cost of diffusion-based models is extremely high due to the iterative sampling procedure 
\citep{song_denoising_2022,lu_dpm-solver_2022,liu_pseudo_2022}. To ameliorate this issue, advanced samplers and novel modeling frameworks have been proposed. For example, \citet{song_score-based_2021} proposed the probability-flow-ODE sampling strategy which inspires the development of DDIM \citep{song_denoising_2022} and DPM-solver \citep{lu_dpm-solver_2022}. \citet{rombach_high-resolution_2022} and \citet{vahdat_score-based_2021} concurrently propose to map data into a lower-dimensional latent space and use a diffusion model to fit the distribution of latent codes. In the application of image generation, \citet{ho_denoising_2020} demonstrate that DDPM synthesizes images in a progressive manner, \ie, the layout information in the intermediate noise (\eg, shape and color) emerges first while the details are enhanced later. This phenomenon facilitates image editing in the latent noise space, \eg, image interpolation and inpainting. Our work also exploits the progressive generation property to achieve semantic mixing in the latent noise space.

\subsection{Controllable Image Generation} Generative models can be used to synthesize images conditioned on certain control signals \citep{kingma_auto-encoding_2014,goodfellow_generative_2020,oord_pixel_2016,kobyzev_normalizing_2021}, such as class labels, text descriptions \citep{saharia_photorealistic_2022,yu_scaling_2022,ramesh_hierarchical_2022}, and degraded images \citep{kawar_denoising_2022}. Many approaches have been developed based on auto-regressive models, variational auto-encoders (VAE), generative adversarial networks (GAN), and diffusion/score-based models. For example, for text-to-image generation, \citet{yu_scaling_2022} propose to model the probability densities of image tokens conditioned on text tokens in an auto-regressive manner; 
\citet{saharia_photorealistic_2022} directly approximates the conditional probability densities of images in the RGB space with a diffusion model. 
To reduce the computational cost of diffusion-based generation, \citet{rombach_high-resolution_2022} proposed a latent diffusion model that compresses images into lower-dimensional codes and models the conditional distribution of latent codes.

\subsection{Image Editing} Semantic mixing is related to several image editing tasks. The first one is masked image inpainting which aims to fill in the masked region with reasonable contents \citep{lugmayr_repaint_nodate,saharia_image_2021,peng_generating_2021,zhao_uctgan_2020}. Without semantic guidance about the empty area, generative models tend to synthesize contents such that the entire image locates in a high-density region. Users cannot interactively control the synthesized contents to be of interest \citep{lugmayr_repaint_nodate}. Even though certain semantic guidance is given, the generated contents may not look harmonious with other parts of the original image. 

The second related task is style transfer which attempts to transfer the artistic style of one source image to another target one \citep{gatys_neural_2015,karras_style-based_2019,luan_deep_2017,ulyanov_texture_2016,zhu_unpaired_2020} by modifying the color, shape, and texture of the target image in a global manner. However, style transfer cannot change the semantic content of the target image. On the other hand, semantic mixing aims to inject the content semantics from another object into the layout semantics; it automatically detects which part of the layout object is to be modified (\eg, when mixing the camel sign with ``husky'' in Fig.~\ref{fig:teaser}, only the camel is replaced by husky while the overall layout remains unchanged). The resultant image looks natural both entirely and locally. 

The third related task is text-driven image editing based on diffusion generative models. Recent works \citep{hertz_prompt--prompt_2022, gal_image_2022, couairon_diffedit_2022,kawar_imagic_2022,wu_unifying_2022,chandramouli_ldedit_2022,kwon_diffusion-based_2022} explore using diffusion generative models for text-driven image editing, such as object replacement, style or color changes, object additions, \etc\ However, unlike our semantic mixing, such editing does not lead to synthesis of new unknown object/concept, which is the main focus of this work.
Compositional generation, on the other hand, composes multiple individual components to generate a complex scene. For example, \citet{liu_compositional_2022} factorizes the diffusion model conditioned on multiple prompts as the product of diffusion models conditioned on each prompt separately. Thus, it can combine scenes described in multiple prompts into one image.
Different from these tasks, semantic mixing aims to fuse multiple semantics into one single object instead of composing multiple objects within one image.

\begin{figure}
    \centering
    \includegraphics[width=0.9\textwidth]{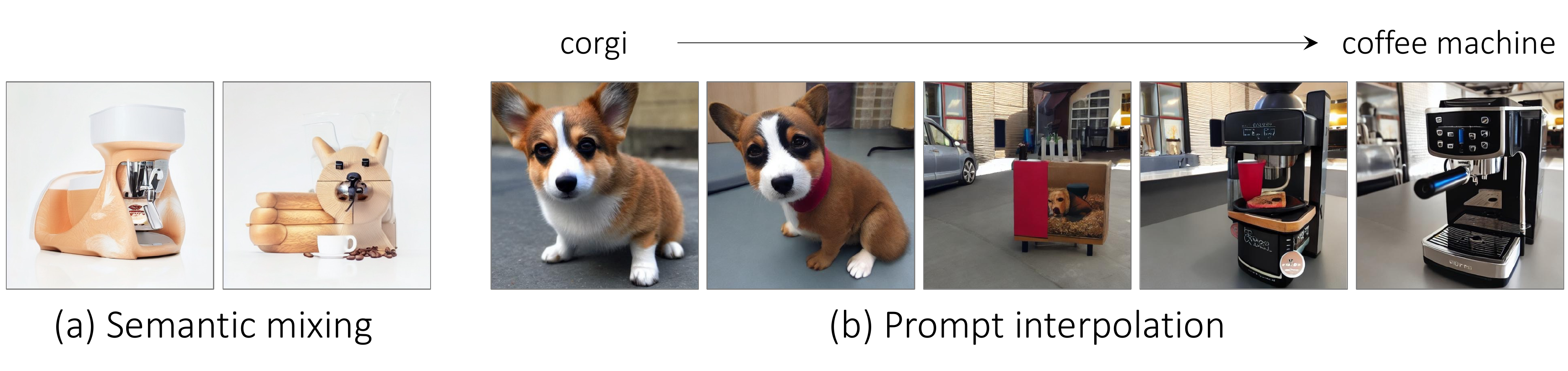}
    \caption{\textbf{Prompt interpolation} fails to yield plausible output when the two concepts are extremely dissimilar (\eg, ``corgi'' and ``coffee machine'').}
    \label{fig:prompt_interpolation}
\end{figure}

Another related task is prompt interpolation, where two different text prompts are interpolated in the text latent space before being used for content generation. However, such approach only works well for prompts with similar semantics (\eg, two dog breeds or two faces). In cases where the two concepts are extremely dissimilar (\eg, ``corgi'' and ``coffee machine''), the generated content is usually dominated by one of the concepts (Figure~\ref{fig:prompt_interpolation}). On the contrary, our semantic mixing can successfully blend two semantics that are highly dissimilar.

\section{Method}

In this section, we first introduce the background of denoising diffusion probabilistic model (DDPM). Then, we formulate the new problem of semantic mixing which intends to combine two different semantics to create novel concepts, and propose an effective diffusion-based framework for implementing such objective. Besides, we discuss two application instances of the proposed framework and elucidate the implementation details.

\subsection{Preliminaries on diffusion models}
Deep generative modeling aims to approximate the probability densities of a set of data via deep neural networks. The deep neural networks are optimized to mimic the distribution from which the training data are sampled \citep{ho_denoising_2020, kingma_auto-encoding_2014, goodfellow_generative_2020, song_score-based_2021}. 
Denoising diffusion probabilistic models (DDPM) are a family of latent generative models that approximate the probability density of training data via the reversed processes of Markovian Gaussian diffusion processes \citep{sohl-dickstein_deep_nodate, ho_denoising_2020}.

Given a set of training data $\sD=\{\vx^i\}_{i=1}^{N}$, for all $i=1,\dots,N$, $\vx^i \in \sR^d$ and are i.i.d. sampled from certain data distribution $ q(\cdot)$, DDPM models the probability density $ q(\vx)$ as the marginal of the joint distribution between $\vx$ and a series of latent variables $x_{1:T}$, 
$$p_{\theta}(\vx)=\int p_{\theta}(\vx_{0:T}) d \vx_{1:T} \quad \text{with} \quad \vx = \vx_0.$$ 
The joint distribution is defined as a Markov chain with learned Gaussian transitions starting from the standard normal distribution $\mcal{N}(\cdot; \vzero, \mI)$, \ie,
$$p_{\theta}(\vx_T)=\mcal{N}(\vx_T; \vzero, \mI), \quad 
p_{\theta}(\vx_{t-1}|\vx_{t})\equiv \mcal N(\vx_{t-1}; \vmu_{\theta}(\vx_t, t), \Sigma_{\theta}(\vx_t, t)).$$
and thus
$$p_{\theta}(\vx_{0:T}) = p_{\theta}(\vx_T)\prod_{t=T}^{1}p_{\theta}(\vx_{t-1}|\vx_{t}).$$

To perform likelihood maximization of the parameterized marginal $\ptheta(\cdot)$, DDPM uses a fixed Markov Gaussian diffusion process, $q(\vx_{1:T}|\vx_0)$, to approximate the posterior $\ptheta(\vx_{1:T}|\vx_0)$. In specific, two series, $\alpha_{0:T}$ and $\sigma^2_{0:T}$, are defined, where $1=\alpha_0 > \alpha_1 > \dots, >\alpha_T \geq0$ and $0=\sigma^2_0 < \sigma^2_1 < \dots < \sigma^2_T$. For any $t>s\geq0$, 
$$q(\vx_{t}|\vx_{s})=\mathcal{N}(\vx_{t}; \alpha_{t|s}\vx_{s}, \sigma^2_{t|s}\mI), \quad \text{where} \quad \alpha_{t|s} = \frac{\alpha_t}{\alpha_s}, \quad \sigma^2_{t|s} = \sigma^2_{t} - \alpha^2_{t|s} \sigma^2_{s}.$$
Thus, 
$$q_(\vx_t|\vx_0) = \mathcal{N}(\vx_t|\alpha_t \vx_0, \sigma^2_t \mI).$$ 
The parameterized reversed process $\ptheta$ of DDPM is optimized by maximizing the associated evidence lower bound (ELBO):
\begin{align}
    -\log \ptheta(\vx_0) & \leq -\log \ptheta(\vx_0) + \KL(q(\vx_{1:T}|\vx_0) \| \ptheta(\vx_{1:T}|x_0)) \nonumber \\
    & = \KL(q(\vx_T|\vx_0)\| \ptheta(\vx_T)) + \sum_{t=1}^{T} \KL(q(\vx_{t-1}|\vx_{t}, \vx_0)\|\ptheta(\vx_{t-1}|\vx_{t})) \nonumber
\end{align}

Given a well-trained DDPM, $\ptheta(\cdot)$, we can generate novel data via various types of samplers, including the Langevin ancestral sampling and probability-flow ODE solvers \citep{song_score-based_2021}. During the reversed process (sampling procedure), a signal with random Gaussian noise will be progressively transformed into a data point located on the manifold of training data. In the case of image generation, an image with pure noise will gradually evolve into a semantically meaningful and perceptually high-quality image. In each stage, we can estimate the true clean image from the corresponding noise, and the reconstructions develop from coarse to fine \citep{ho_denoising_2020}. More specifically, it has been shown that the sampling procedure of DDPMs first crafts the layouts or profiles of the final output images, and then, synthesizes the details such as the face of a human or the texture of a flower. Consider a certain intermediate step where the noise has already contained the information of layout, \citet{ho_denoising_2020} demonstrate that if we fix the noise and run multiple sampling actions starting from this step, the resultant images will share the common layout. Inspired by this phenomenon (\textbf{progressive generation}), we will explore how to use diffusion-based models to perform semantic mixing, \ie, given a certain semantic layout, can we mix it with any arbitrary contents of our interests?

\begin{figure}[t!]
    \centering
    \includegraphics[width=0.9\linewidth]{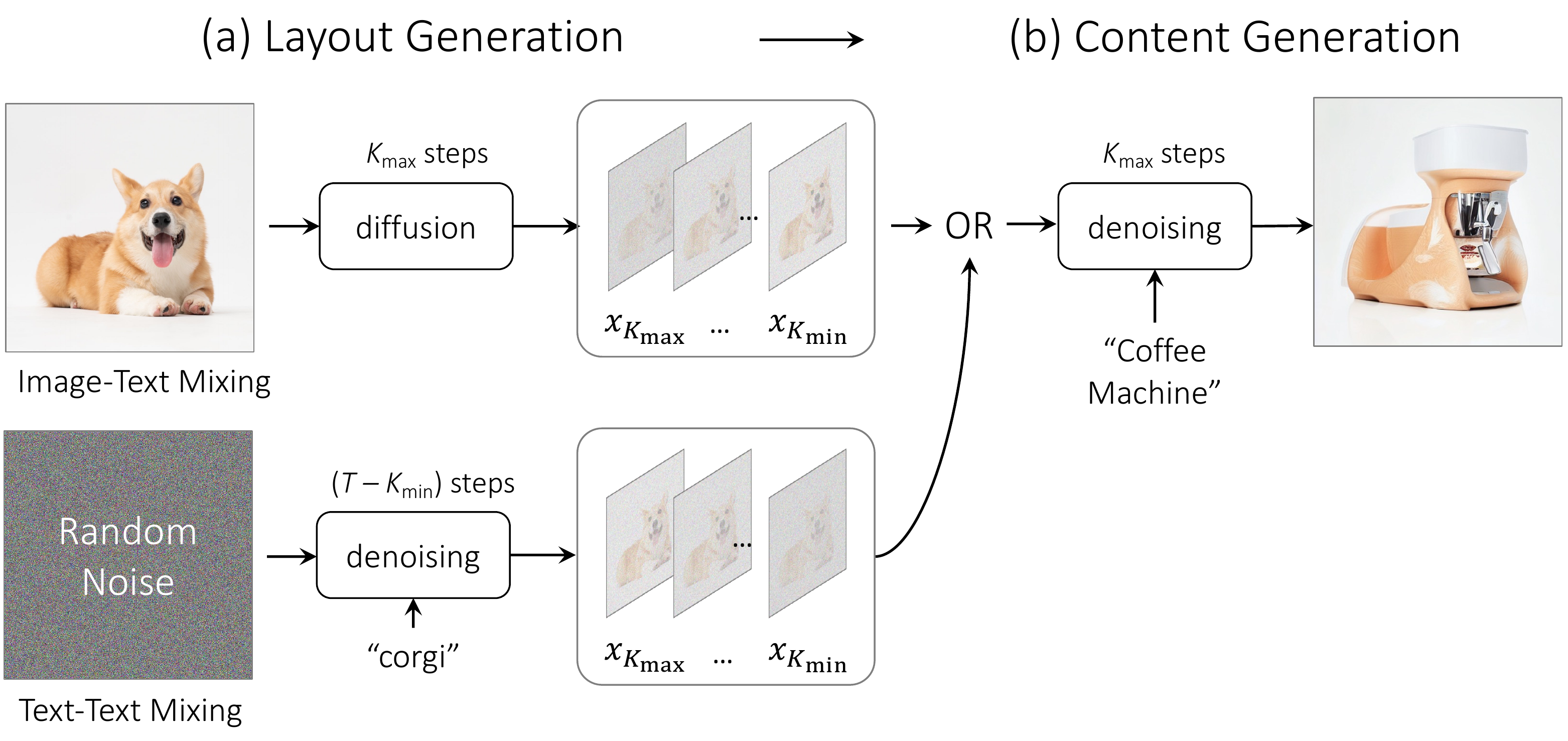}
    \caption{\textbf{The overall pipeline of \nameofproject.} \nameofproject\ is built upon pre-trained text-to-image diffusion-based generative models. It enables semantic mixing of two different concepts (\eg, ``corgi'' and ``coffee machine'') by first synthesizing coarse layout semantics (either by adding Gaussian noise to a given image or denoising from a random Gaussian noise given a text prompt), followed by denoising on condition of the desired concept (``coffee machine'' in this example) to obtain a new concept (\eg, corgi-alike coffee machine). Image credit (input image): Unsplash.}
    \label{fig:framework}
\end{figure}

\subsection{Semantic mixing with diffusion models}
The creation of new concepts and objects plays an important part in multimedia productions, such as creating anthropomorphic animation roles. One paradigm of concept creation is to mix the semantics of multiple things. For example, many classical animation characters are designed by mixing animal faces with human bodies, such as the ``Monkey King'' and ``Puss in Boots''. 
In this section, we introduce a novel task of image generation, \textbf{{semantic mixing}}, which aims to modify the \textit{content} in a certain part of a given object while preserving its \textit{layout} semantics. The new content is synthesized based on the content semantics of another object. 
For example, given the shape and color layout semantic extracted from one object (\eg, a watermelon slice), one can generate an object of certain content semantic (\eg, a lamp) in that shape and color.

Inspired by the progressive generation property of diffusion-based models, we propose a method, \textbf{{\nameofproject}}, to blend the semantics of two objects. \nameofproject\ utilizes a pre-trained text-to-image diffusion-based generative model, $\ptheta(\vx|\vy)$, to extract and mix two semantics. The overall framework is illustrated in figure \ref{fig:framework}.
The layout semantic can be extracted from either a given image or a text prompt, while the content semantic is determined by a conditioning text prompt. We can generate images of mixed semantics by denoising the noisy layout images with a conditioning content prompt. Depending on the input type for layout generation, our \nameofproject\ can operate in two different modes: (a) image-text mixing and (b) text-text mixing.

\begin{figure}
    \centering
    \includegraphics[width=\linewidth]{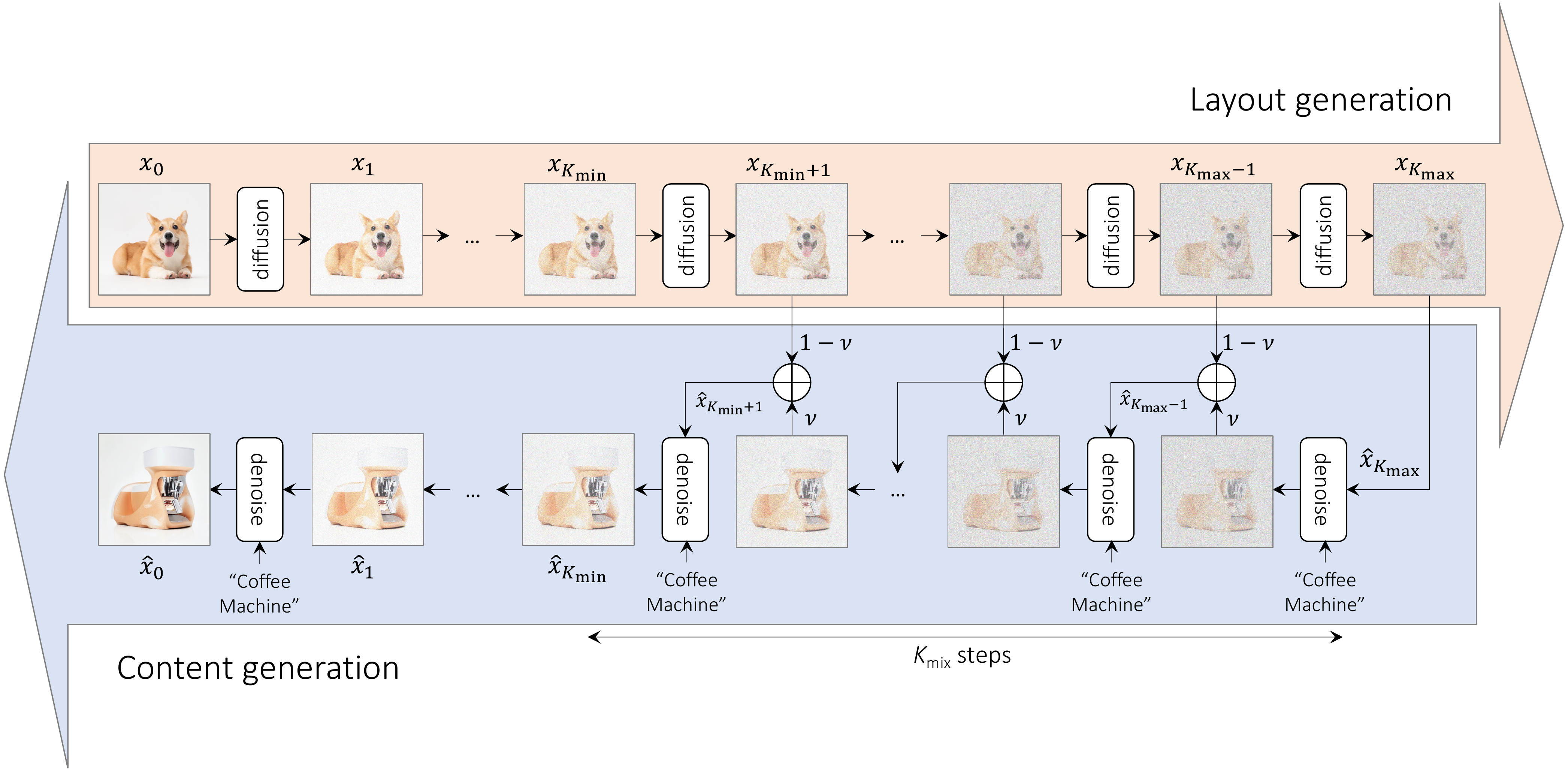}
    \caption{\textbf{The detailed pipeline of \nameofproject\ (image-text mixing).} Given an image $\vx_0$ of layout semantics, we first craft its corresponding layout noises from step $K_{\text{min}}$ to $K_{\text{max}}$. Starting from $K_{\text{max}}$, the conditional generation process progressively mixes the two concepts by denoising given the conditioning content semantics (``coffee machine'' in this example). For each step $k \in [K_{\text{min}}, K_{\text{max}}]$, the generated noise of mixed semantics is interpolated with the layout noise $\vx_k$ to preserve more layout details. For $k \in [0, K_{\text{min}}]$, no interpolation is used.}
    \label{fig:detailed_pipeline}
\end{figure}

\smallskip
\noindent \textbf{(a) Image-text mixing.}
In the case where the layout semantic is specified by a given image $\vx$, we first generate its noisy versions corresponding to the intermediate steps from $K_{\text{min}}$ to $K_{\text{max}}$. Each of the noisy images $\{\vx_{k}\}_{K_{\text{min}}}^{K_{\text{max}}}$ consists of the layout and profile information of the given image $\vx$, with coarse-to-fine layout.
Then, we perform the denoising process by conditioning the text of the content semantic $\vy$. The reversed process starts from the noise of layout semantic $\hat{\vx}_{K_{\text{max}}} = \vx_{K_{\text{max}}}$. For each step $k$ from $K_{\text{max}}$ to $K_{\text{min}}$, the denoising process utilizes information from the generative model $\ptheta(\hat{\vx}_{k-1}|\hat{\vx}_k, y)$ as well as the information from the layout noise $\vx_{k-1}$. In specific, we first sample $\hat{\vx}'_{k-1} $ from $\ptheta(\hat{\vx}'_{k-1}|\hat{\vx}_k, y)$. Then, we perform a linear combination of $\vx_{k-1}$ and $\hat{\vx}'_{k-1}$ with a constant $\nu \in [0,1]$ to craft the mixed noise $\hat{\vx}_{k-1} = \nu \hat{\vx}'_{k-1} + (1-\nu)\vx_{k-1}$. From step $K_{\text{min}}$ to $0$, the denoising process only depends on the conditional generative model and no linear interpolation is applied. Figure~\ref{fig:detailed_pipeline} illustrates the detailed process of image-text mixing.

\smallskip
\noindent \textbf{(b) Text-text mixing.}
In the other case where the layout semantic is determined by a text prompt $y_{\text{layout}}$, we first sample a sequence of layout noise $\{\vx_k\}_{K_{\text{min}}}^{K_{\text{max}}}$ from the distribution $\ptheta(\vx_k|\vy_{\text{layout}})$. Then, similar to the case of image-text mixing, we iteratively denoise the layout noise to synthesize image of mixed semantics via the generation process conditioned on $\vy_{\text{content}}$. Interpolations are still only applied from step $K_{\text{max}}$ to $K_{\text{min}}$.

\subsection{Controlling the mixing ratio}
While being able to synthesize images with mixed semantics, it remains unclear how to control the amount of blending elements, \eg, increasing the element of ``coffee machine'' or preserving more elements of ``corgi''. 
Next, we present several tricks to provide better control and flexibility over the generated content.

\begin{figure*}[b]
    \centering
    \includegraphics[width=\textwidth]{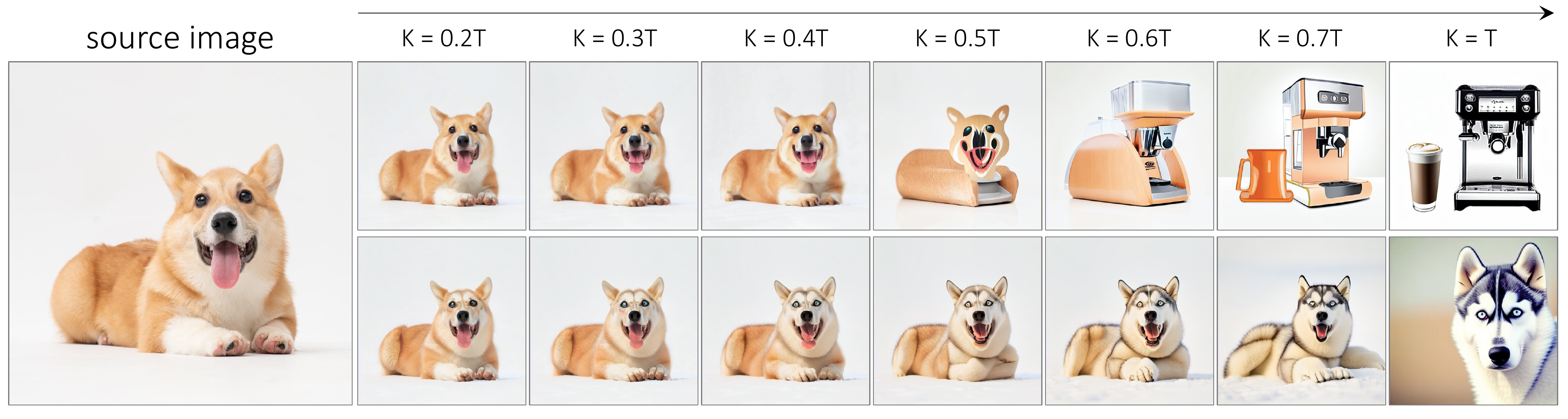}
    \caption{\textbf{Varying time-step for content injection.}
    Consider $K=K_{\text{max}}=K_{\text{min}}$. In this example, ``corgi'' is blended with ``coffee machine'' and ``husky'' in the top and bottom row, respectively.
    The time-step to inject the content conditioning affects the preservation of the layout semantics and the fusion of the content semantics.
    Image credit (source image): Unsplash.}
    \label{fig:timestep_injection}
\end{figure*}

\subsubsection{Time-step for injecting content prompt}
As discussed earlier, \nameofproject\ enables mixing of two different concepts by first crafting noisy images of layout semantic from step $K_{\text{max}}$ to $K_{\text{min}}$, followed by injection of conditional prompt. We choose $K_{\text{min}}$ such that the noisy layout image contains rich details from the given layout image and choose $K_{\text{max}}$ such that the irrelevant details are destroyed and only a coarse layout is preserved. By integrating noise across different time steps, the generation process can inject content semantics to proper regions in the given layout image and preserve more layout semantics such as shape and color.

\noindent \textbf{Varying time-step for content injection.}
In Figure~\ref{fig:timestep_injection}, let $K = K_{\rm max} = K_{\rm min}$ and $\nu=1$, we first study the effect of varying the time-step $K$ for content injection. We first notice that when $K$ is small, the generation process $\ptheta(\hat{\vx}_{0}|\vx_{K}, \vy_{\text{content}})$ can only modify a small part of image content due to limited number of denoising steps available. As a result, we can fuse two concepts with similar semantics (\eg, corgi and husky) but fail to mix two very different objects (\eg, corgi and coffee machine). For example, with $K=0.4T$, the eyes and texture of husky begin to appear on the face of corgi but no elements of ``coffee machine'' is found when mixing ``corgi'' with ``husky'' and ``coffee machine'', respectively. On the other hand, to enable mixing of two distinct objects, the generation process conditioned on $\vy_{\text{content}}$ requires much larger $K$ to ensure sufficient steps for mixing. As shown in the top row of Figure~\ref{fig:timestep_injection}, given $K=0.6T$, the conditional generation process successfully synthesizes a coffee machine in the shape of a corgi. 

\begin{figure}
    \centering
    \includegraphics[width=\textwidth]{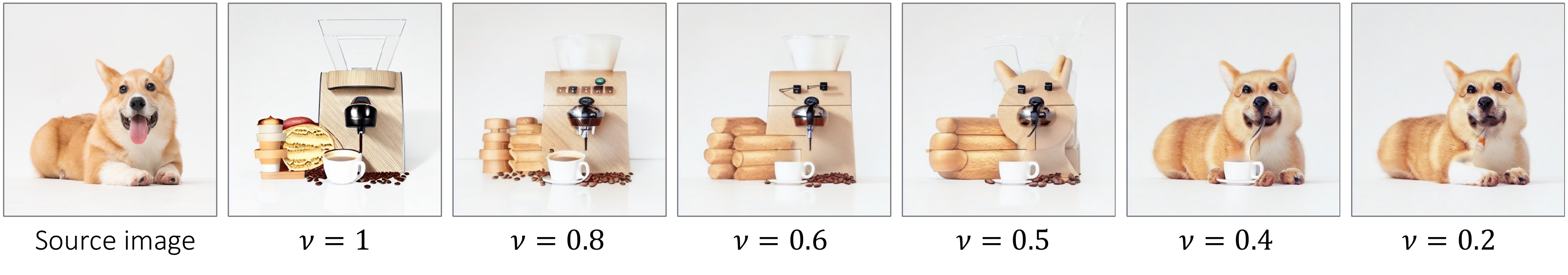}
    \caption{\textbf{Linear interpolation between layout noise and conditionally generated noise}. The constant $\nu$ controls the ratio between the layout (``corgi'') and content semantics (``coffee machine''). Image credit (source image): Unsplash.}
    \label{fig:method_nu}
\end{figure}

\noindent \textbf{Preserving more layout details.}
To preserve more elements of the given layout object, we perform denoising starting from step $\vx_{K_{\text{max}}}$ and propose to interpolate the original layout noise with the synthesized noise obtained from the conditional generation process earlier. The mixing constant $\nu$ controls the ratio between layout and content semantics. Once again, we show an example of mixing the layout of a ``corgi'' with the content of a ``coffee machine'' in Figure~\ref{fig:method_nu}. When $\nu=1$, the conditional generation process starts from step ${K_{\text{max}}}$ and uses no information from $\{\vx_k\}_{K_{\text{min}}}^{K_{\text{max}}}$. We can synthesize an image of ``coffee machine'' in a similar color as the ``corgi'' image but contains almost no element of ``corgi'' asides from the shape.
Interesting, when $\nu\leq 0.4$, we notice that only a coffee cup is synthesized due to the dominance of ``corgi'' elements. 
In this example, we can obtain an image of corgi-alike coffee machine when we set $\nu=0.5$ or $0.6$.
In practice, we fix $K_{\rm max}=0.6T$ and $K_{\rm min}=0.3T$, and vary only $\nu$.

\noindent \textbf{Optimal value of $\nu$.}
We also notice that the ``optimal'' interpolation constant $\nu$ is determined by the semantic similarity between the two concepts. 
For example, when mixing ``corgi'' and ``husky'', diffusion models only need to modify the eyes and texture. Therefore, we can use a small value of $\nu$ (\eg, $0.1$).
On the contrary, when mixing ``corgi'' and ``coffee machine'', since the two concepts are extremely dissimilar, the diffusion models require more denoising steps in order to overwrite the rabbit details. In this case, we can use a large value of $\nu$ (\eg, $0.9$).

\subsubsection{Weighted image-text cross-attention}\label{sec:cross_attn}
\begin{figure}
    \centering
    \includegraphics[width=\textwidth]{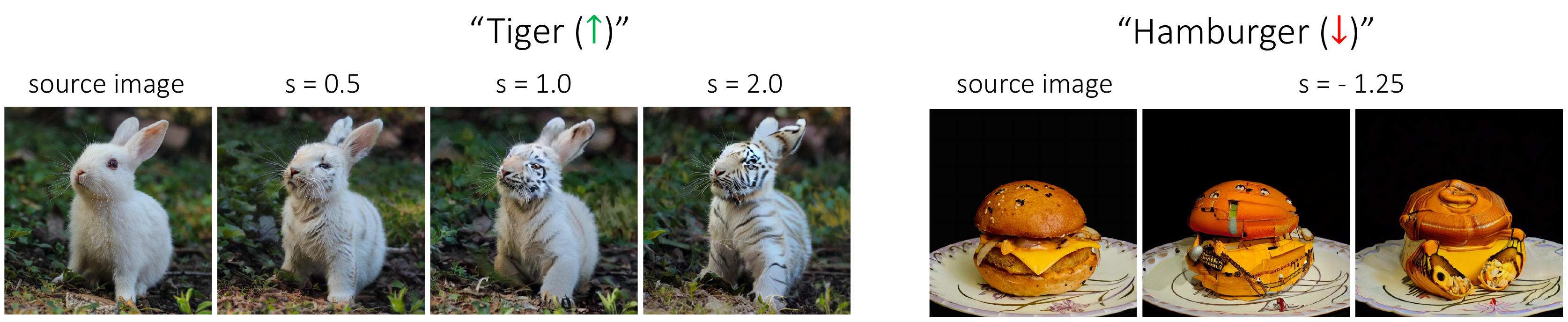}
    \caption{\textbf{Image-text cross-attention re-weighting.} Scaling the attention map of the desired words enables adjustment of the corresponding elements (\eg, observe the change in amount of tiger stripes given different $s$). On the contrary, multiplying the cross-attention map by a negative scale $s$ removes the concept, forcing the model to generate a new object that does not belong to the original class.  }
    \label{fig:cross_attn_strength}
\end{figure}

Inspired by Prompt-to-Prompt~\citep{hertz_prompt--prompt_2022}, we also find it effective to re-weigh the image-text cross-attention to increase or reduce the magnitude of a concept. Consider the case of mixing ``rabbit'' and ``tiger''. Given text-image cross-attention maps $\mM \in \mathcal{R}^{N_{\rm image} \times N_{\rm text}}$, where $N_{\rm image}$ and $N_{\rm text}$ denote number of spatial and text tokens, respectively, and a conditional prompt $\vy = $``a photo of tiger'', we scale the attention map corresponding to the ``tiger'' tokens with parameter $s \in [-2, 2]$ while keeping the remaining attention maps unchanged. As shown in Figure~\ref{fig:cross_attn_strength}, the extent of ``tiger'' content can be adjusted using different values of positive scale $s$ (\eg, the amount of tiger stripes).

\noindent \textbf{Concept removal.}
On the other hand, we observe that applying negative $s$ leads to an interesting behaviour: given a hamburger image and a conditioning prompt $\vy = $``a photo of hamburger'', using a negative $s$ amounts to encouraging the diffusion models to generate an image with a layout similar to that of hamburger while not being a hamburger. We call this \textbf{concept removal}. As shown in the right subfigure of Figure~\ref{fig:cross_attn_strength}, when ``hamburger'' concept is eliminated, the diffusion model is forced to imagine the most probable non-hamburger object, such as an airship or crab.

\subsection{Implementation details}
In practice, we use Latent Diffusion Models (LDM) for semantic mixing. Since the auto-encoder in LDM is trained with patch-wise losses, the auto-encoder preserves the spatial correspondence between the latent space and the original RGB space. We also observe the progressive generation property in the sampling procedure of LDMs. Our implementation is developed based on the
Stable Diffusion\footnote{Stable-Diffusion: \url{https://huggingface.co/CompVis/stable-diffusion}} code base which is an open-source implementation of LDM. One can use Stable Diffusion to generate high-quality images. It also offers multiple types of samplers to balance the trade-off between sample quality and computational efficiency. We use DDIM sampler in our experiments. 
\section{Applications}
\label{sec:application}
In this section, we show several applications using our \nameofproject, including (a) semantic style transfer (Section~\ref{sec:semantic_style_transfer}), (b) novel object synthesis (Section~\ref{sec:novel_object_synthesis}), (c) breed mixing (Section~\ref{sec:breed_blending}) and (d) concept removal (Section~\ref{sec:concept_removal}). 

\subsection{Semantic style transfer} \label{sec:semantic_style_transfer}
\begin{figure}
    \centering
    \includegraphics[width=\textwidth]{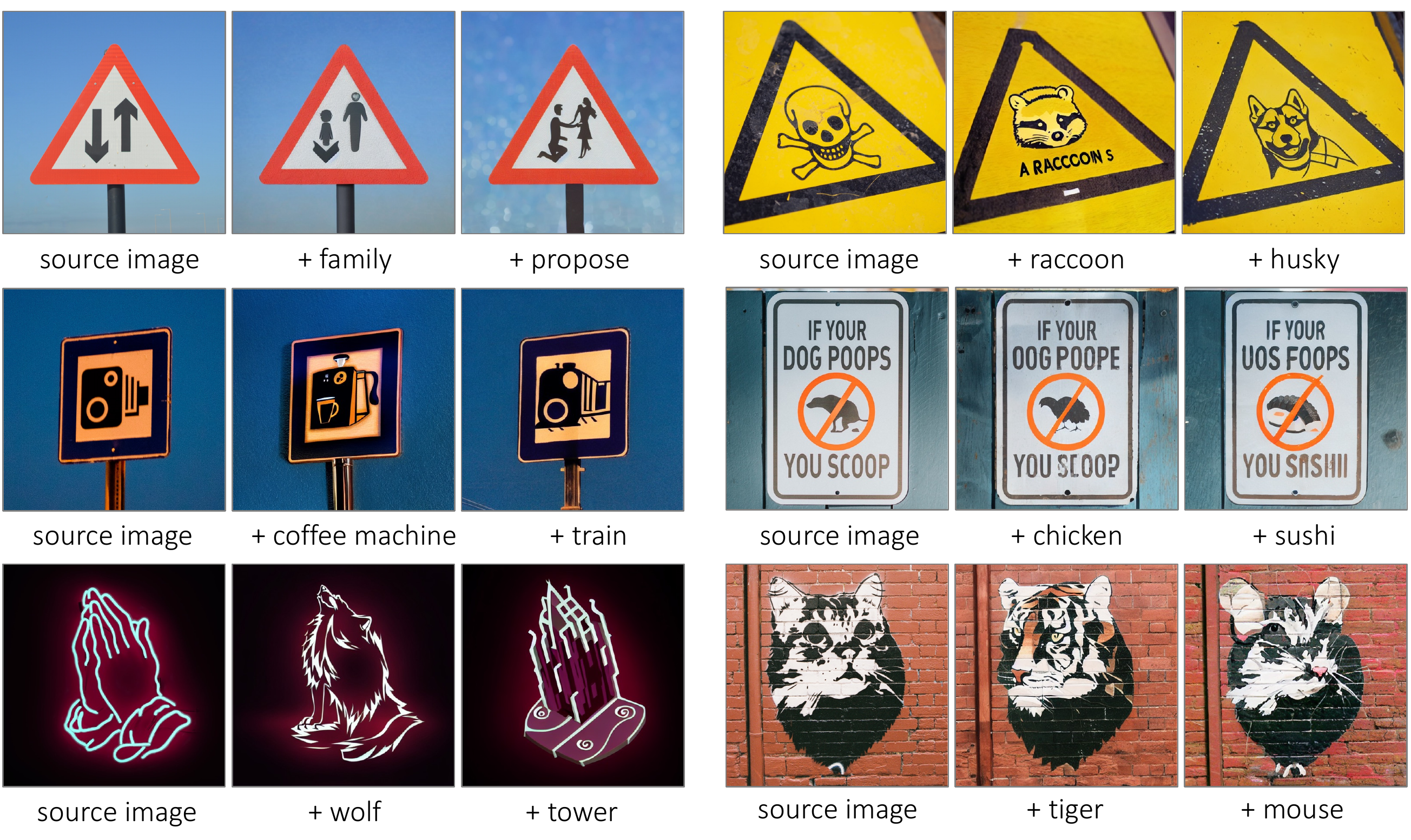}
    \caption{\textbf{Semantic style transfer.} A new sign can be synthesized by injecting new concept-of-interest (\eg, replacing the skull with raccoon or husky). Note that the spatial layout and background is well preserved despite the large content change. Image credit (source images): Unsplash.}
    \label{fig:semantic_style_transfer}
\end{figure}

We first demonstrate semantic style transfer application by synthesizing signs with different semantics (\eg, replacing the arrows in two-way sign with people). 
Unlike style transfer where the content image is stylized based on the reference style image without changing the image content, our \nameofproject\ allows the user to inject new semantics while preserving the spatial layout and geometry (\eg, the triangular sign). We show some examples in Figure~\ref{fig:semantic_style_transfer}. Note that the background is well-preserved despite the large content change. 
Such application could possibly be used to assist design of new logo/sign by injecting a new concept to a template.

\subsection{Novel Object Synthesis} \label{sec:novel_object_synthesis}
\begin{figure*}
    \centering
    \includegraphics[width=\textwidth]{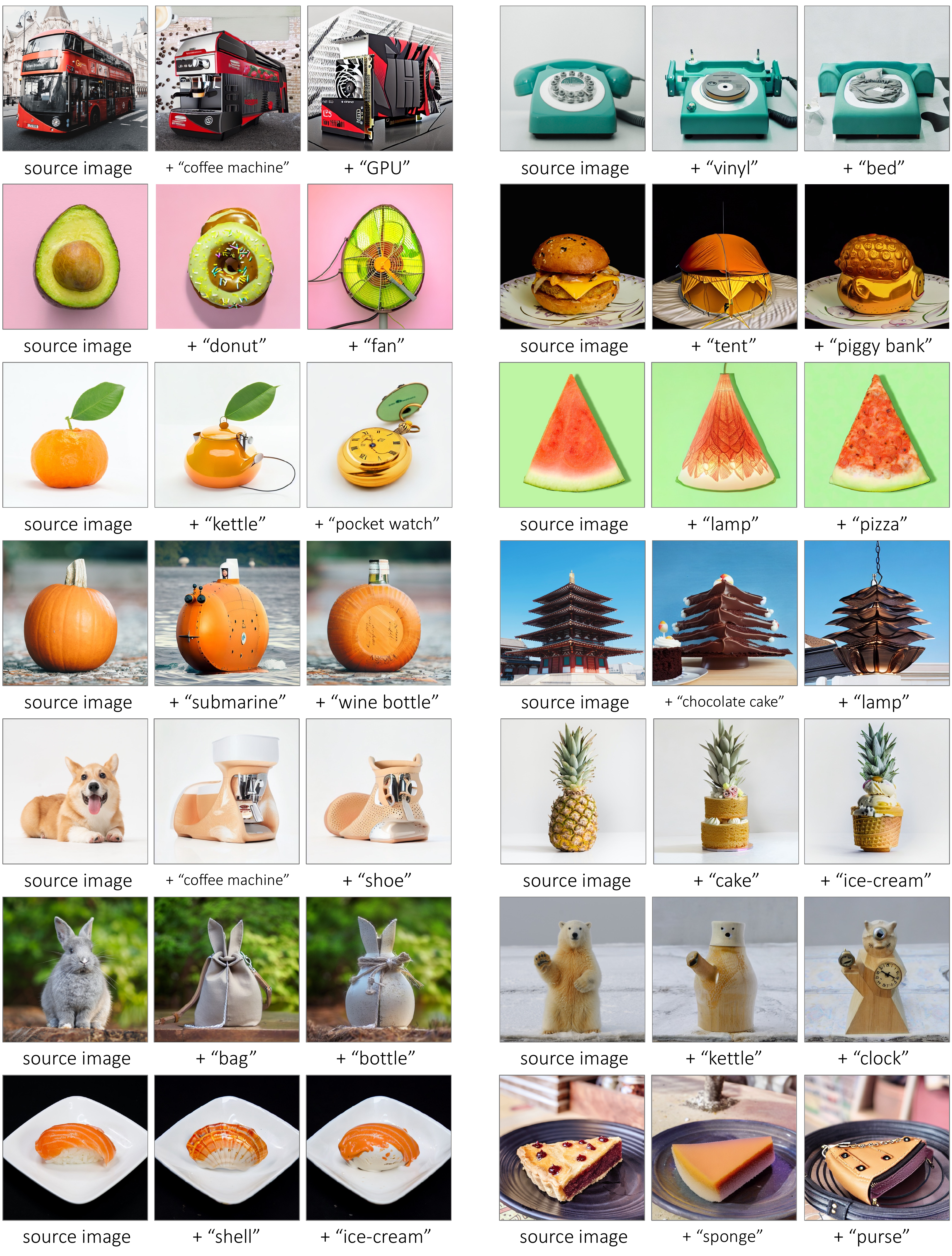}
    \caption{\textbf{Novel object synthesis} enables creation of new object by injecting a new concept into existing object (\eg, bus $+$ coffee machine). Note that the image background adapts to the conditioning prompt (\eg, road has turned into sea when ``submarine'' is injected to the pumpkin image.). Image credit (source images): Unsplash.}
    \label{fig:novel_object_synthesis}
\end{figure*}

Our \nameofproject\ also allows the synthesis of novel objects by injecting new concepts (\eg, coffee machine) into an existing object (\eg, bus). 
This can be extremely useful in inspiring creativity when designing new commercial products.
We show some examples in Figure~\ref{fig:novel_object_synthesis}. It is worth nothing that the background context adapts accordingly based on the conditioning prompt. For example, the road has turned into sea when the ``submarine'' is mixed with a pumpkin image.
Similarly, when a pagoda is mixed with ``chocolate cake'', the road has become a table to better fit the entire image context.
This suggests that mixing of the two concepts occurs at a semantic level.

\subsection{Breed mixing} \label{sec:breed_blending}
\begin{figure*}
    \centering
    \includegraphics[width=\textwidth]{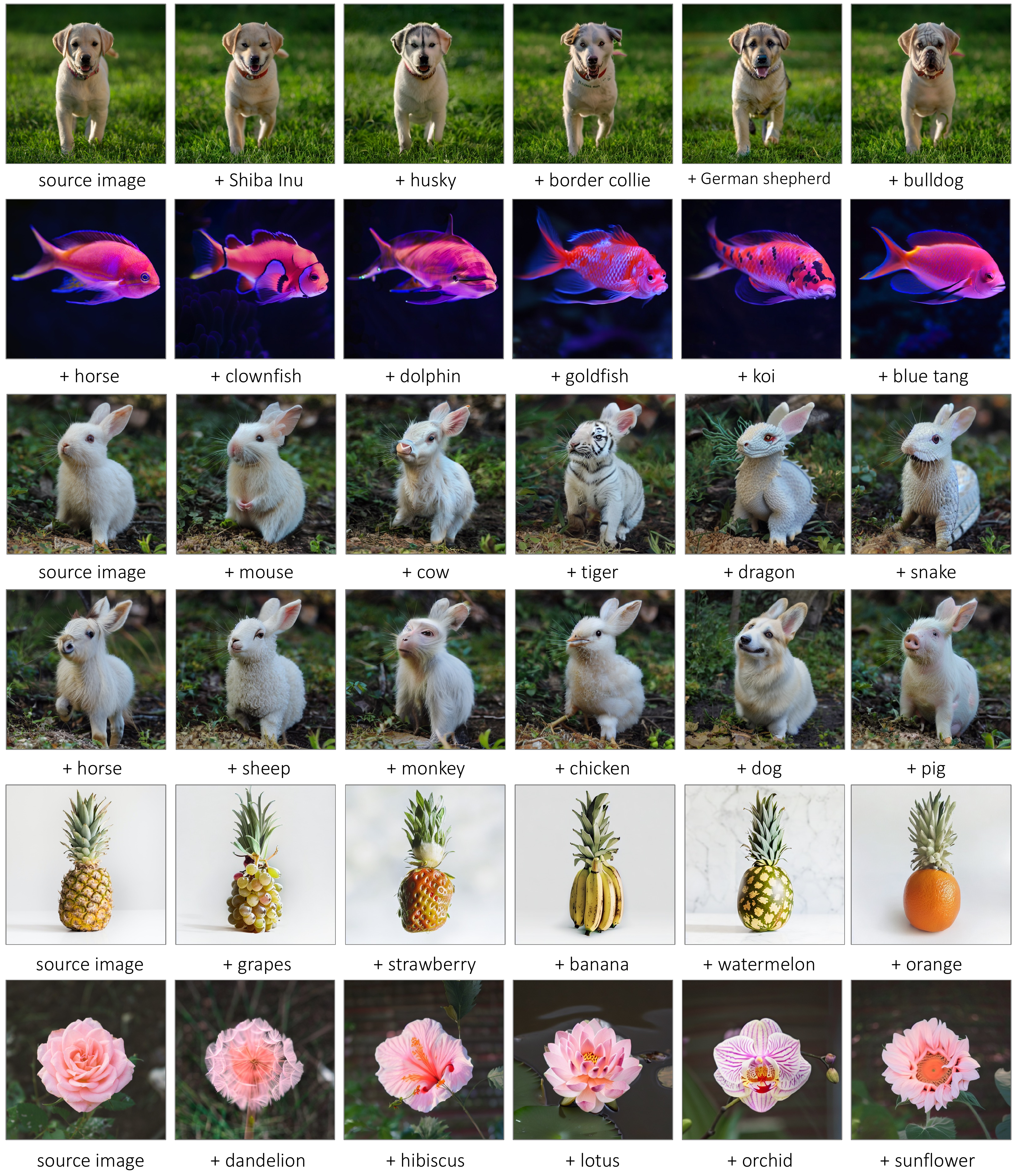}
    \caption{\textbf{Breed mixing.} (1st-2nd rows) Our method enables mixing of two different animal breeds (\eg, Labrador and bulldog) while retaining each species' distinct features (Labrador's ears and bulldog's face). (3rd-4th rows) In addition, our \nameofproject\ allows mixing of two different animal species and synthesizes new unseen species (\eg, rabbit + tiger) with high-quality synthesis and fidelity. (5th-6th rows) Similarly, we can also mix two different fruits (\eg, pineapple and grapes) or flower species (\eg, rose and dandelion) to create a new species. Image credit (source images): Unsplash.}
    \label{fig:breed_blending}
\end{figure*}

Next, we demonstrate the possibility of mixing two different breeds given our method. As depicted in the first two rows of Figure~\ref{fig:breed_blending}, our method can mix two different animal breeds (\eg, Labrador and bulldog) and generate plausible results with distinct features (Labrador's ears and bulldog's face). More interestingly, our method can even mix two different species and generate new unseen animal species as depicted in the third and forth rows.
Note that some of these combinations share almost no commonalities (\eg, rabbit and chicken, rabbit and tiger), yet we can still obtain photo-realistic results. Similarly, in the last two rows, we also demonstrate the mixing of two different fruits (\eg, pineapple and grapes) or flowers (\eg, rose and dandelion).

\subsection{Concept Removal} \label{sec:concept_removal}
\begin{figure*}
    \centering
    \includegraphics[width=\textwidth]{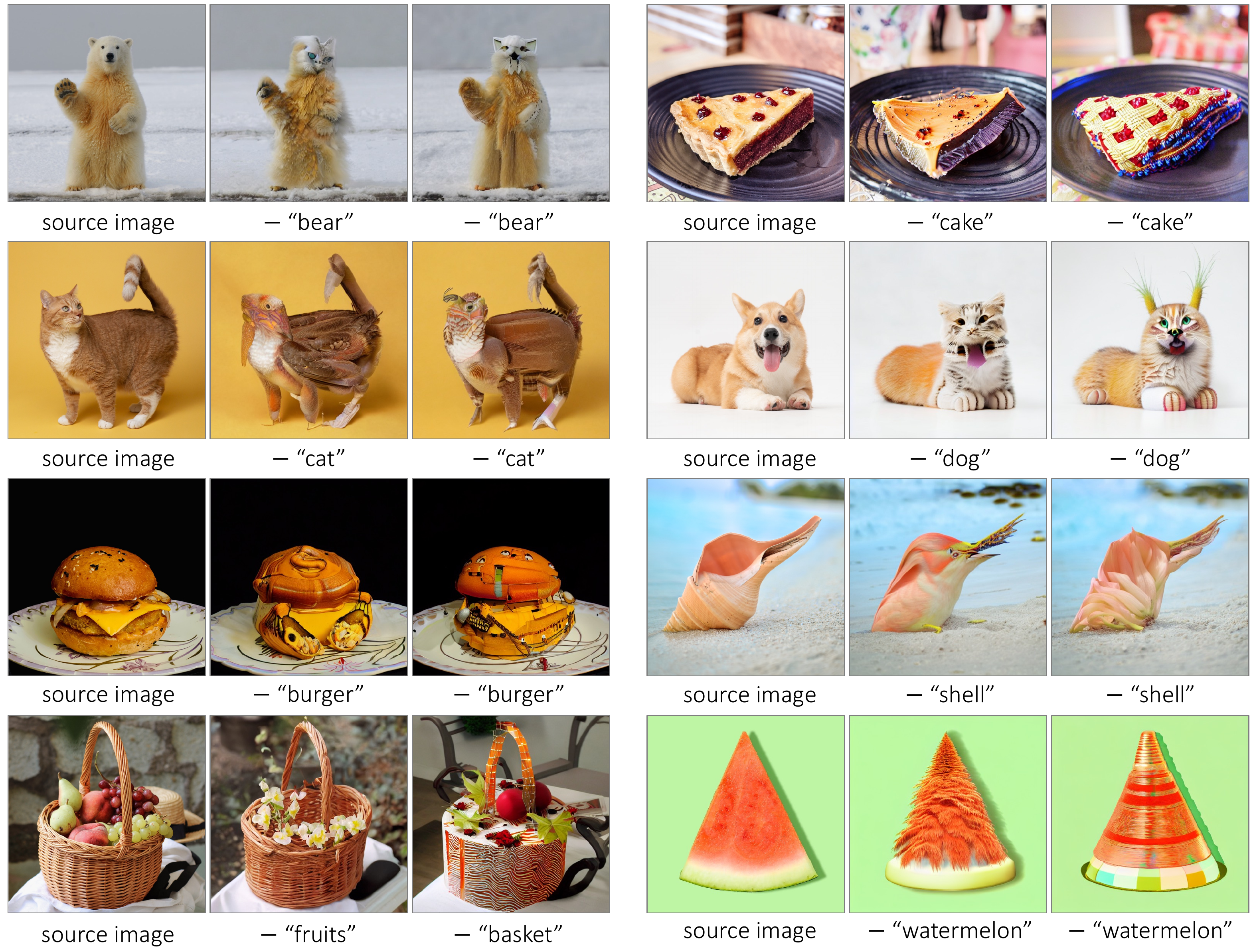}
    \caption{\textbf{Concept removal} enables synthesis of new image without its original content. Note that the overall layout is largely preserved while the original semantic is removed. Image credit (source images): Unsplash.}
    \label{fig:concept_removal}
\end{figure*}

We have presented various applications by injecting new semantics into existing ones. Here, we are also interested in generating a new image by removing its original semantic and let the model to decide what to generate aside from its original content. This can be easily achieved by multiplying the image-text cross-attention map by a negative weight (Section~\ref{sec:cross_attn}). Some examples are shown in Figure~\ref{fig:concept_removal}. As we can see, the generated images largely preserve the overall layout while removing the original semantics. For example, as shown in the last row, given a basket of fruits, by removing the concept ``fruits'', we obtain a basket of flowers instead. On the other hand, removing the concept ``basket'' leads to generation of a cake with fruits on top.

\subsection{Text-text Semantic Mixing}
\begin{figure}
  \centering
  \includegraphics[width=1.\textwidth]{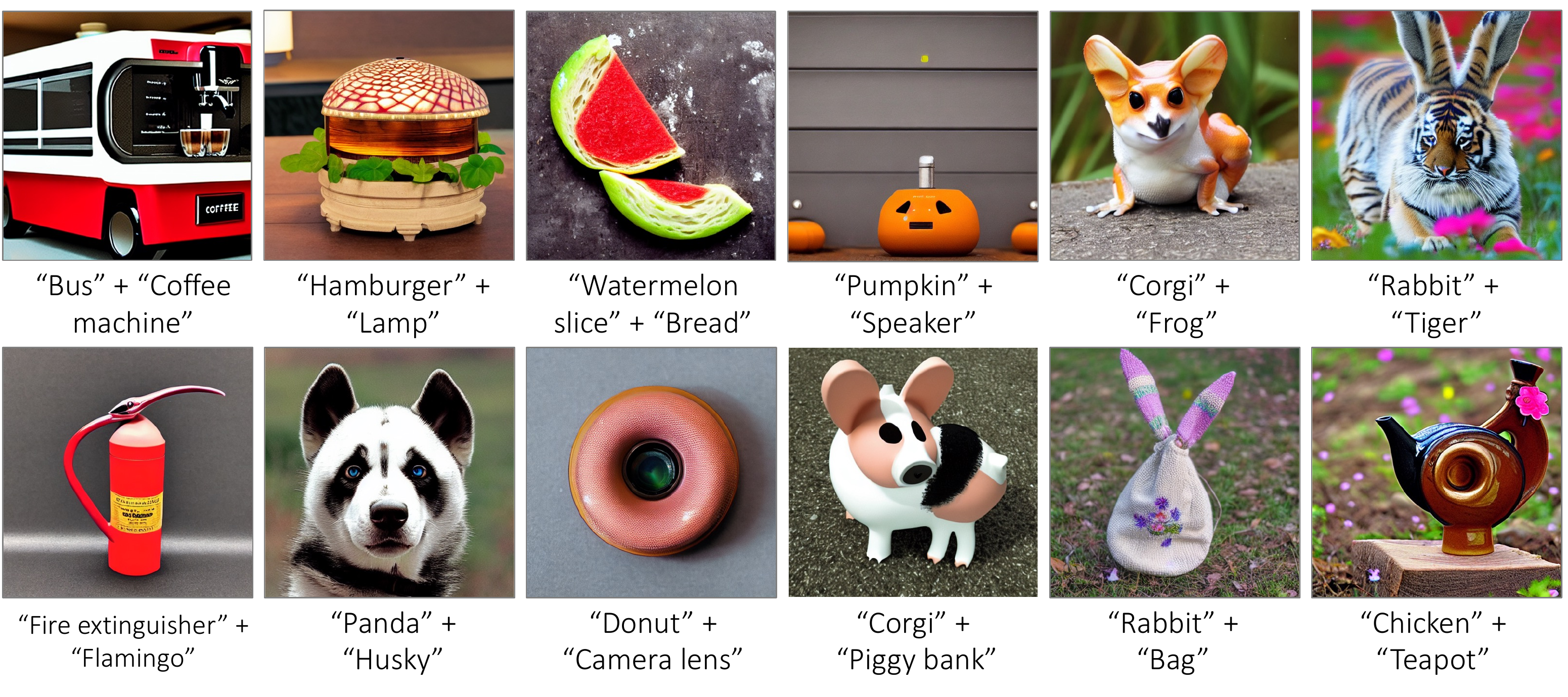}
  \caption{\textbf{Examples of text-text semantic mixing} where no input image is required.}
  \label{fig:text-text_blending}
\end{figure}

In the previous sections, we have demonstrated several applications of our \nameofproject\ using image-text mixing (layout semantics is crafted based on a given image). Next, we also provide some results of \nameofproject\ using text-text mixing mode where no image is needed. As shown in Figure~\ref{fig:text-text_blending}, our method successfully mixes two different semantics and generates photo-realistic results. However, one limitation of text-text semantic mixing is that, since no image is provided for layout generation, the final synthesis result is unpredictable.

\section{Limitations}
\begin{figure}
    \centering
    \includegraphics[width=\textwidth]{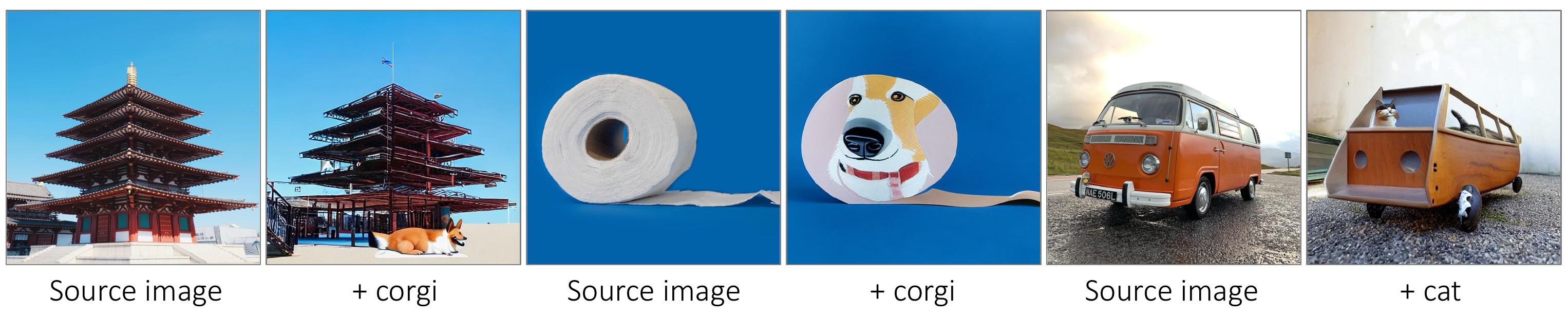}
    \caption{\textbf{Failure cases}. When the two concepts do not share any shape similarity, \nameofproject\ will instead compose the two (\eg, a cat riding a van or a painting of corgi on the toilet roll).}
    \label{fig:limitation}
\end{figure}

We identify a failure case of our method where two concepts cannot be mixed if they do not share any shape similarity (\eg, mixing ``van'' and ``cat'' or ``toilet roll'' and ``corgi''). In this case, the two concepts will be simply composited (\eg, a cat riding a van or a painting of corgi on the toilet roll). Some examples can be found in Figure~\ref{fig:limitation}. We leave solving these to future work. 

\section{Conclusion}
In this work, we present a novel task called \textbf{semantic mixing}, whose objective is to mix two different semantics to synthesize a new unseen concept. 
To this end, we present \textbf{\nameofproject}, a simple solution based on pre-trained text-conditioned diffusion-based image generation models. Our method exploits the properties of diffusion-based generative models by injecting new concepts during the denoising process. Our approach does not require any spatial masks or re-training, while preserving the layout and geometry. Given this, our \nameofproject\ supports several downstream applications, including semantic style transfer, novel object synthesis, breed mixing and concept removal. 

\section{Societal Impact}
The goal of our work is to synthesize a novel object of mixed concepts. Similar to other deep learning-based image synthesis and editing algorithms, our method has both positive and negative societal impacts depending on the applications and usages. On the positive side, \nameofproject\ could inspire creation of new commercial products (\eg, corgi-alike coffee machine). On the down side, it could be used by malicious parties to deceive or mislead humans.
Another issue is that the pre-trained model used in this work, Stable Diffusion v1.4~\citep{rombach_high-resolution_2022}, was trained on LAION dataset, which is known to have social and cultural bias.

\bibliographystyle{samples/iclr2020_conference}
\bibliography{samples/_references}

\end{document}